\documentclass[runningheads]{llncs}
\usepackage[T1]{fontenc}
\usepackage{graphicx}
\usepackage[dvipsnames,table]{xcolor}
\usepackage{multirow}
\usepackage{array}
\usepackage{acronym}
\usepackage{tikz}
\usepackage[ruled,vlined]{algorithm2e}
\usepackage{amsmath}
\usepackage{amssymb}
\usepackage{booktabs}
\usepackage{makecell}
\usepackage{amsfonts} 
\usepackage{hyperref}

\usepackage{color,soul}

\definecolor{CellBck}{gray}{0.92}
\definecolor{CellBlue}{RGB}{240, 240, 255}
\definecolor{OracleColor}{gray}{0.5}
\definecolor{AugColor}{RGB}{25, 175, 25}
\definecolor{DecrColor}{RGB}{175, 25, 25}

\definecolor{OracleColor}{RGB}{100, 100, 100}
\definecolor{Methodcolor}{gray}{0.1}

\definecolor{AlgCommentColor}{RGB}{31, 153, 46}

\newcommand{\cbg}{\cellcolor{CellBck}}

\definecolor{MethodColor}{gray}{0.4}
\definecolor{CellBck}{gray}{0.92}

\acrodef{MLP}[MLP]{multi-layer perceptron}
\acrodef{MRD}[MRD]{measurable residual disease}
\acrodef{AML}[AML]{acute myeloid leukemia}
\acrodef{ALL}[ALL]{acute lymphoblastic leukemia}
\acrodef{FCM}[FCM]{flow cyotmetry}
\acrodef{DL}[DL]{deep learning}
\acrodef{CNN}[CNN]{convolutional neural networks}
\acrodef{SOTA}[SOTA]{state-of-the-art}
\acrodef{GNN}[GNN]{graph neural network}
\acrodef{GMM}[GMM]{gaussian mixture models}
\acrodef{b-ALL}[b-ALL]{b-cell acute lymphoblastic leukemia}
\acrodef{ST}[ST]{set transformer}
\acrodef{FPS}[FPS]{farthest-point-sampling}
\acrodef{GAT}[GAT]{graph attention network}
\acrodef{GCN}[GCN]{graph convolution network}
\acrodef{ISAB}[ISAB]{induced set attention block}
\acrodef{GIN}[GIN]{graph isomorphism network}
\acrodef{ASAP}[ASAP]{adaptive structure aware pooling}
\acrodef{PMA}[PMA]{pooled multi-head attention}

\newcommand{\method}[1]{\texttt{\textbf{\color{MethodColor} #1}}}

\begin{document}
\title{On the importance of local and global feature learning for automated measurable residual disease detection in flow cytometry data}
\titlerunning{Local and global feature learning for FCM data}
%
\author{Lisa Weijler\inst{1} \and
Michael Reiter\inst{1,2} \and
Pedro Hermosilla\inst{1} \and
Margarita Maurer-Granofszky\inst{2}\and
Michael Dworzak\inst{2}}
\authorrunning{L. Weijler et al.}
%
\institute{TU Wien \and
St.Anna CCRI\\
}
\maketitle              

\begin{abstract}
This paper evaluates various deep learning methods for measurable residual disease (MRD) detection in flow cytometry (FCM) data, addressing questions regarding the benefits of modeling long-range dependencies, methods of obtaining global information, and the importance of learning local features. Based on our findings, we propose two adaptations to the current state-of-the-art (SOTA) model. Our contributions include an enhanced SOTA model, demonstrating superior performance on publicly available datasets and improved generalization across laboratories, as well as valuable insights for the FCM community, guiding future DL architecture designs for FCM data analysis. 
The code is available at \url{https://github.com/lisaweijler/flowNetworks}.

\end{abstract}
\section{Introduction}
\label{sec:intro}
The detection and monitoring of \ac{MRD} in pediatric acute leukemia represent a critical aspect of patient care and treatment evaluation~\cite{campana2010minimal}. \Ac{MRD} defined as the proportion of residual cancer cells in patients after therapy, serves as a prognostic indicator for disease relapse and guides therapeutic decisions towards achieving better clinical outcomes~\cite{dworzak2008standardization,testi2019outcome}. \Ac{FCM}, with its ability to analyze cellular characteristics at a single-cell level, has emerged as a cornerstone technique for \ac{MRD} assessment due to its sensitivity and specificity~\cite{mckinnon2018flow}.

However, the accurate identification and quantification of \ac{MRD} amidst heterogeneous cell populations remain challenging, often necessitating complex data analysis methodologies and training of medical experts. In recent years, the advent of \ac{DL} approaches has revolutionized the landscape of biomedical data analysis~\cite{dash2020deep}, offering promising solutions to address the inherent complexities of \ac{MRD} detection in \ac{FCM} data for pediatric acute leukemia.

Given the unstructured characteristic of single cell \ac{FCM} data, it does not fit in well-researched modalities such as text or images, and hence, applying \ac{DL} methods is not straightforward. While traditional machine learning approaches have become standard practice for the analysis of \ac{FCM} data~\cite{suffian2022machine,cheung2021currenttrends,hu2022application}, there have only been a handful of approaches applying \ac{DL} to \ac{FCM} data directly, primarily relying on \ac{CNN} for e.g. imaging \ac{FCM} or attention-based networks that can process unstructured data for single cell \ac{FCM} data.
 
\ac{FCM} data samples are essentially sets of $F$-dim feature vectors (events) corresponding to single cells in a feature space $\mathbb{R}^F$ comprised of the properties measured by \ac{FCM}, where $F$ is usually between $10$ and $15$ and can vary between different samples. Similar cell types share similar feature vectors and tend to form clusters, representing a composition of different cell populations.

Since events within one sample are not i.i.d., previous works suggest that the relative position of cell populations within one sample contains crucial information for successful \ac{MRD} detection, meaning it is beneficial to process whole samples at once rather than considering single events detached from their origin-sample as input to \ac{DL} models~\cite{rota2016role,reiter2019automated,licandro2018wgan}. In other words, global feature extraction is suggested to be beneficial for single-cell classification as in the task of \ac{MRD} detection. However, to the best of our knowledge, no extensive evaluation of this assumption exists except for minor baseline testing with simple MLPs. Further, there are several ways of learning global features and infusing them with single-cell features for classification, which have not been evaluated. Dominating methods in the literature rely on \ac{GMM} or self-attention, while the latter holds the current \ac{SOTA} for automated \ac{MRD} in pediatric \ac{b-ALL}~\cite{wodlinger2021automated}. Self-attention allows for modeling long-range dependencies yet does not explicitly learn local features, i.e., introduce an inductive bias of spatial locality, which is a crucial component of common successful architectures using convolutions and local feature aggregation, especially for tasks requiring fine semantic perception~\cite{guo2020deep}.

In this work, we provide an extensive evaluation of several methods guided by asking the following questions.
"\textit{Does automated \ac{MRD} detection benefit from modelling long-range dependencies?}", "\textit{Does it matter how the global information is obtained?}" and "\textit{Is it beneficial to explicitly learn local features?}" 
Based on the analysis findings, we propose two adaptations of the current \ac{SOTA} that give a performance increase and better generalization abilities. First, the current \ac{SOTA} is based on the \ac{ST}~\cite{lee2019set}, which uses learned query vectors, called inducing points, to mitigate the quadratic complexity issue of self-attention; instead of using learned query vectors we propose to use feature vectors from the \ac{FCM} sample directly sampled by \ac{FPS}. Second, we introduce explicit learning of local features by infusing the self-attention layers with \ac{GNN} layers.

In summary, our main contributions are,
\begin{itemize}
    \item[1.] an enhanced version of the current \ac{SOTA} model that leads to a new \ac{SOTA} performance on publicly available datasets as well as better generalization abilities between datasets of different laboratories,
    
    \item [2.] providing an extensive evaluation of several \ac{DL} methods for \ac{FCM} data with valuable insights for the \ac{FCM} community, on which future designs of \ac{DL} architectures can be based.
\end{itemize}

\section{Related Work}

Given the wide range of applications of \ac{FCM} data, the developed approaches for automated analysis are highly task-specific. In this work, we focus on single-cell classification for rare cell populations and give an overview of techniques related to this task. 

The most direct approach to automatically predicting the class label of each event of a patient sample is to pool events from the training set of different samples together and train a classifier using pairs of single events and corresponding labels.
Authors in~\cite{abdelaal2019LDA} propose a linear discriminant analysis classifier (LDA) for this task, leading to interpretable results due to the simplicity of LDA. However, the assumption of equal covariance matrices of classes is not valid for rare cell population detection as in \ac{MRD} quantification. In~\cite{Ni2016SVM}, authors suggest training one support vector machine model per patient, implicitly incorporating prior knowledge of the patient's specific phenotype. However, this requires the availability of labeled training data for each patient. 

Other methods are based on neural networks~\cite{licandro2018wgan}. However, methods using single events as input are restricted to learning fixed decision regions. One way to circumvent this is to register samples by transforming them into a standard feature space~\cite{weijler2020umaprf,li2017gating} or by creating landmarks based on prior biological knowledge that guides classification~\cite{Lee2017ACDC}. Another way is to process whole samples at once, yet those methods need to be equivariant to the ordering and handle the volume of events present in a sample. Representing a sample based on its statistical parameters by, e.g., Gaussian Mixture Models~\cite{reiter2019automated} is an option. Others propose sample-wise clustering-based approaches~\cite{weijler2022umaphdbscan,bruggner2014citrus,Weber2019cdiffcyt}.

Methods based on \ac{DL} that process whole samples at once are scarce, given the characteristics of FCM data. One line of research is to transform FCM samples to images and apply convolutional neural networks as in~\cite{Arvaniti2017cellcnn}. More recently, methods based on the attention mechanism~\cite{vaswani2017attention} have been proposed~\cite{wodlinger2021automated,kowarsch2022xai,weijler2024fate}. Attention-based models are a way for event-level classification that learns and incorporates the relevance of other cell populations in a sample for the specific task. A limiting factor is the complexity of the standard self-attention operation that increases quadratically with the input sequence length $\mathcal{O}(n^2)$; this is infeasible for \ac{FCM} data and efficient variants such as the \ac{ST}~\cite{lee2019set} have to be used, as authors in~\cite{wodlinger2021automated} do.

In the context of \ac{FCM} data, to the best of our knowledge, graph-based methods have only been applied for unsupervised clustering, where the clustering algorithm itself is graph-based~\cite{becht2019dimensionality,Levine2015phenograph}, but not for targeted cell classification and modeling local spatial structure. Additionally, as far as we know, this work is the first to introduce the benefits of \ac{GNN} to use sample- or patient-specific features implicitly.

\section{Methods}
\label{sec:methods}
In this section, the problem setup is described in detail (Section~\ref{subsec:pre}), all methods analyzed are introduced (Section~\ref{subsec:arch}), and the experimental setup is outlined (Section~\ref{subs:exp_setup}).

\subsection{Preleminaries}
\label{subsec:pre}
We treat the problem of \ac{MRD} detection as a binary classification of single events into healthy or cancerous cells. Throughout the paper, we use the following definitions of \ac{FCM} data sets.
\begin{definition}
\label{def:FCMdataset}
A \ac{FCM} data set $\mathcal{X} = \{X_1,\dots,X_N\}$ contains $N$ samples $X_i \in \mathbb{R}^{n_i \times F_i}, i= 1\dots N$, where $F_i$ is the feature space dimension of sample $X_i$ and $n_i =|X_i|$ the number of events $x_{i_j}, j=1,\dots, n_i$ per sample. 
\end{definition}

\begin{figure}[t]
  \centering
   \includegraphics[width=1.0\linewidth]{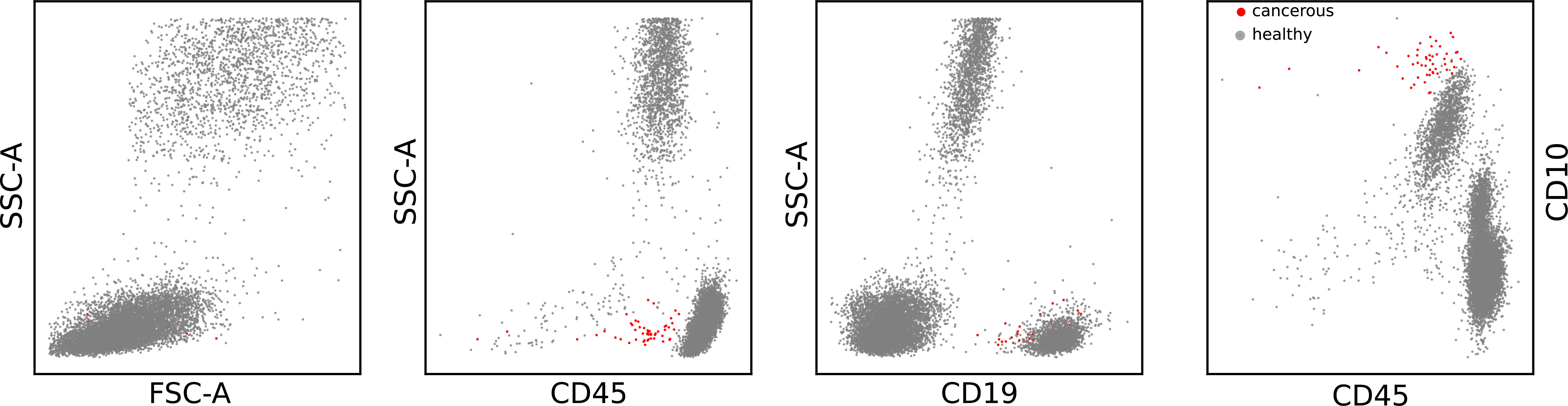}
   \caption{This figure shows 2D projections of an \ac{FCM} sample on pairs of features, where each dot represents the feature vector of a cell (event). Healthy cells are denoted in grey, and cancerous cells in red. FSC, SSC stands for forward-, side-scatter, and CD for cluster of differentiation.}
   \label{fig:panelplot}
\end{figure}

The proportion of leukemic cells varies between samples; it can be as low as $0.01\%$, while the number of measured cells is up to $n_i=10^6$. Fig. \ref{fig:panelplot} shows 2D projections of an \ac{FCM} sample.

When applying \ac{GNN} to \ac{FCM} data, each sample $X_i$ is converted into a graph $\mathcal{G}_i$ by the $k$-NN algorithm utilizing its entire individual feature space $\mathbb{R}^{F_i}$. The graph is constructed using all events $x_{i_{j}}\in X_i$; every graph node thus represents a single event. 

\subsection{Methods assessed}
\label{subsec:arch}
We introduce the architectures analyzed to answer the posed questions in the following. The range of models used is not exhaustive, yet yields a good representation of various domains ranging from simple \ac{MLP} over \ac{GNN} to variations of attention-based networks.

We distinguish between \textit{no-context} models, where cells are classified based on the information of the single cell only, \textit{global-context} models, where the sample as a whole is included in the prediction of single cells, and \textit{local-context} models, where the local context of similar cells in the sample is used for classification. Finally, we present our proposed architecture, \textit{local-global-context}, where explicit extraction of local features is combined with long-range dependency modeling.

All architectures comprise four layers of the specific network layers and a linear layer as prediction-head using a hidden dimensionality of $32$ unless stated otherwise. If multi-head self-attention is part of the architecture, four heads are used. A $k$-NN graph is constructed with $k=10$ for methods that use local feature learning unless stated otherwise. As non-linearity, the GELU activation function is employed. 

\subsubsection{No-context model}
A baseline model for comparing to single cell processing with neural networks, which translates into fixed decision boundaries independent of in-sample-context.
 \paragraph{\method{MLP}:} A simple MLP with batch normalizations after the non-linearity.
\subsubsection{Global-context models}
Baseline models and proposed \ac{FCM}-specific adaptations to compare different versions of incorporating global information, i.e., taking the whole input sample into consideration for single cell predictions.

 \paragraph{\method{MLP-mean}:}  Same as \method{MLP} but a global feature vector obtained with mean-aggregation is infused by concatenation to the single-event feature vectors before passing through the last linear layer
 
 \paragraph{\method{MLP-max}:} Same as \method{MLP-mean} but using max-aggregation. 
 
 \paragraph{\method{MLP-pma}:} Same as \method{MLP-mean} but using a learnt query vector for aggregation. This aggregation is equivalent to the \ac{PMA} proposed in~\cite{lee2019set} using one seed vector and one attention head, where a learned query vector is cross-attended to all single-event feature vectors.
 
 \paragraph{\method{PointNet}:} The PointNet architecture introduced in~\cite{qi2017pointnet} for 3D point cloud classification and segmentation.
 
\paragraph{\method{PointNet-adapted}:} Same as \method{PointNet} but to reweigh the focus on the global information to be equally distributed among event-wise and global vectors, we increase the dimensionality of the event-wise feature vectors to match the dimensionality of the global vectors, namely 1024. 

\paragraph{\method{ST}:} The \ac{ST}~\cite{lee2019set} which holds the current \ac{SOTA} for automated \ac{MRD} detection in \ac{b-ALL}~\cite{wodlinger2021automated}. It uses learned query vectors, called \textit{inducing} points, to circumvent the quadratic complexity of self-attention. The \ac{ISAB} is defined as

\begin{equation}
    \begin{split}
        ISAB_m(X)&=MAB(X,H)\in \mathbb{R}^{n\times d},\\
        \text{where } H&=MAB(I,X) \in \mathbb{R}^{m\times d},
    \end{split}
\end{equation}
with $I\in \mathbb{R}^{m\times d}$ being the $m$ $d$-dimensional inducing points, $X\in \mathbb{R}^{n\times d}$ the set to be processed with cardinality $n$ and $MAB$ a multi-head attention block. Such an \ac{ISAB} reduces the complexity from $\mathcal{O}(n^2)$ to $\mathcal{O}(mn)$ with $m<<n$. Intuitively, \ac{ISAB} summarizes the sample in the learned queries and induces the information back with cross-attention. $m=16$ as proposed in \cite{wodlinger2021automated}.

\paragraph{\method{reluFormer}:} An adaptation of the cosFormer proposed in \cite{qin2022cosformer} that uses ReLU instead of softmax to remove the non-linearity in the attention calculation to be able to rearrange the matrix multiplications to get linear complexity of self-attention while using the entire input sequence length. Authors in \cite{qin2022cosformer} propose a cosine-based distance re-weighting scheme instead of the softmax function that focuses the attention values on neighboring tokens. Since we do not have an ordered sequence as input, we omit the re-weighting scheme for this work. 

\paragraph{\method{ST-FPS}:} Inspired by the intuition of \ac{ISAB}, we propose to select event feature vectors instead of learning queries to compose the matrix as inducing points. We use \ac{FPS}, a widely used sampling method, on the input \ac{FCM} sample to select the indices of events used to create the $I$ matrix. We use a sampling ratio of $r=0.0005$, which results on average in $\approx 150$ events.

\subsubsection{Local-context models}
Baseline models that introduce an inductive bias based on prior knowledge of homophily (biologically similar events share similar feature measurements) by explicitly learning local features.

\paragraph{\method{GCN}:}The \ac{GCN} layers as proposed in \cite{kipf2016semi}. 

\paragraph{\method{GAT}:}The \ac{GAT} layers as proposed in \cite{velickovic2018gat}. 

\paragraph{\method{GIN}:}The \ac{GIN} layers as proposed in \cite{xu2018powerful}. 

\paragraph{\method{GAT-AsAP}:}The \ac{ASAP} layers as proposed in \cite{ranjan2020asap} combined with \ac{GAT} layers. The architecture used is similar to~\cite{ranjan2020asap} with two \ac{GAT} blocks, constituted of 2 \ac{GAT} layers each and two \ac{ASAP} layers pooling to 100 and 50 nodes, respectively. 

\paragraph{\method{GIN-AsAP}:}Same as \method{GAT-ASAP} but with a \ac{GIN} layer instead of \ac{GAT}.

\subsubsection{Local- and global-context model}
The proposed architecture combines explicit local feature learning with long-range dependency modeling. 
\paragraph{\method{GAT-ST-FPS}:}The adapted \ac{SOTA} architecture based on the findings in Section \ref{sec:experiments}, \ac{GAT} infused \ac{ST} with \ac{FPS} self-attention. We use one \ac{GAT} layer for local feature extraction and concatenate those with the input feature vectors before inserting them into the three \method{ST-FPS} layer.

\begin{figure}[t]
  \centering
   \includegraphics[width=1.0\linewidth]{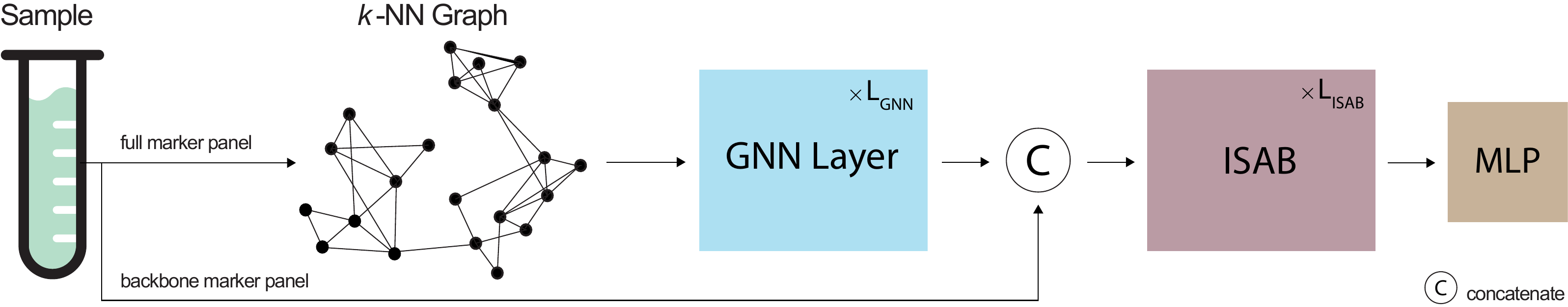}
   \caption{This figure shows the general architecture of the proposed local- and global-context model. For our experiments, we used one \ac{GNN} layer and three \ac{ISAB} with \ac{FPS} instead of learned query vectors. The prediction head \ac{MLP} is a linear layer.}
   \label{fig:method}
\end{figure}

\paragraph{\method{GIN-ST-FPS}:}\ac{GIN} infused \ac{ST} with \ac{FPS} self-attention. It is the same as \method{GAT-ST-FPS} but with a \ac{GIN} layer instead of \ac{GAT}.
Figure~\ref{fig:method} shows an overview of the proposed architecture.

\subsection{Experimental setup}
\label{subs:exp_setup}
\paragraph{Datasets.}
We conduct our experiments on publicly available\footnote{\href{https://flowrepository.org/id/FR-FCM-ZYVT}{flowrepository.org}} data sets of bone marrow samples from pediatric patients with \ac{b-ALL}. In our main experiments, we use the most extensive dataset \textit{Vie}; for our inter-laboratory experiments to assess the architectures' ability to generalize to different laboratories, we use \textit{Bln} and \textit{Bue} for testing. 

\begin{itemize}
    \item \textit{Vie} contains 519 samples collected between 2009 and 2020 at the St. Anna Children's Cancer Research Institute (CCRI) with an LSR II flow cytometer (Becton Dickinson, San Jose, CA) and FACSDiva v6.2. The samples collected between 2009 and 2014 were stained using a conventional seven-color drop-in panel ("B7") consisting of the liquid fluorescent reagents: CD20- FITC/ CD10-PE/ CD45-PerCP/ CD34-PE-Cy7/ CD19-APC/ CD38-Alexa-Fluor700 and SYTO 41. The samples collected between 2016 and 2020 were stained using dried format tubes (DuraClone™, "ReALB") consisting of the fluorochrome-conjugated antibodies CD58-FITC/ CD34-ECD/ CD10-PC5.5/ CD19-PC7/ CD38-APC-Alexa700/ CD20-APC-Alexa750/ CD45-Krome Orange plus drop-in SYTO 41.
    \item \textit{Bue} contains 65 samples collected between 2016 and 2017 at the Garrahan Hospital in Buenos Aires. The samples were recorded with FACSCanto II flow cytometer with FACSDiva v8.0.1 and stained with the following panel: CD58, FITC/CD10, PE/CD34, PerCPCy5.5/CD19, PC7/CD38, APC/CD20, APC-Alexa750/CD45, Krome-Orange plus drop-in SYTO 41.
    \item \textit{Bln} consists of 72 samples collected in 2016 at the Charité Berlin. The samples were collected with a Navios flow cytometer and stained with the same panel as \textit{Bue}.
\end{itemize}

All data were collected on day 15 after induction therapy. Local Ethics Committees approved sampling and research, and informed consent was obtained from patients or patient's parents or legal guardians according to the Declaration of Helsinki. Ground truth was obtained using manual gating by at least two experts. For every sample, the resulting labels from different experts are combined into a final gating for each sample to obtain reliable ground truth data.

Table~\ref{tab: data} provides a tabular overview of the data sets.

\begin{table}
    \centering
    \caption{Description of the FCM data sets.}
    \label{tab: data}
    \begin{tabular}{ cccc }
        \toprule
        {\bf Name} & {\bf City} & {\bf Years} & {\bf Samples}\\ [0.5ex] 
        \hline 
        {Vie} & Vienna & 2009-2020 & 519 \\ 
        {Bln} & Berlin & 2016 & 72 \\ 
        {Bue} & Buenos Aires & 2016-2017 & 65 \\
        \bottomrule
    \end{tabular}
\end{table}

\paragraph{Metrics.} We use precision $p$, recall $r$, and $F_1$-score for evaluation, with correctly identified cancer cells as true positives. The metrics are computed per \ac{FCM} sample and then averaged to obtain the final score. We report each experiment's mean and standard deviation of at least five runs.

\paragraph{Training details.} For comparability, we keep the same training setup for each experiment. We use a batch size of 4 with $5*10^4$ randomly sampled events per sample. For augmentation, we employ random jitter with a scale parameter of $0.01$ and label smoothing with $eps= 0.1$. We use a train, validation, and test split of $50\%$, $25\%$, and $25\%$, respectively. Each model is trained for $150$ epochs using AdamW optimization with a learning rate of $0.001$ and cosine annealing with a starting value of
$0.001$, a minimum value of $0.0002$, and a maximum of $10$ iterations as learning rate scheduler. To mitigate overfitting in \ac{GAT} layers, we employ dropout and weight decay with a rate of $0.2$. Training for all experiments is conducted on an NVIDIA GeForce RTX 3090. The model that performed best on the validation split based on mean $F_1$-score is used for testing.

\section{Experiments}
\label{sec:experiments}
In this section, we analyze the results of the methods described above and aim to find answers to the questions posed. First, in Section \ref{subsec:global}, the importance of global information for the success of the current \ac{SOTA} \method{ST}~\cite{wodlinger2021automated} is assessed. Second, in Section \ref{subsec:globalhow}, different ways of obtaining global features are compared. Third, in Section~\ref{subsec:local_benefits}, methods explicitly learning local features are evaluated. And finally, in Section~\ref{subsec:meth_ours}, we combine the findings into the proposed enhanced architecture. 

\subsection{Does automated \ac{MRD} detection benefit from modelling long-range dependencies?}
\label{subsec:global}

To answer this question, we take the current SOTA method \method{ST} and compare it to its identical architecture but with removed self-attention, which is substituted by summed query, key, and value vectors. This way, we can test for the impact of modeling long-range dependencies without architecture noise. Tabel~\ref{tbl:global} shows that modeling long-range dependencies brings a clear benefit, and the architecture itself has no contribution but rather impairs results for event-wise processing when looking at the simple MLP for comparison. This question can thus be answered with an \textit{yes}. However, there are several ways of obtaining global features, which are discussed in the next section.

\begin{table}
\caption{Results for removing the ability to model long-range dependencies without an architecture change from the \mbox{\ac{SOTA}} model.}
\label{tbl:global}
\setlength{\tabcolsep}{8pt}
\begin{center}
\begin{tabular}{lcccc}
    \toprule
    Method &  $p$ & $r$ & avg $F_1$ & med $F_1$ \\
    \midrule
    \method{ST-No-Att} & 0.7463 & 0.8766 & \cbg 0.771 {\scriptsize $\pm 0.0038$} & 0.8925 {\scriptsize $\pm 0.0036$}\\
    \cmidrule{2-5}
    \method{MLP} & 0.7912	 & 0.8701 & \cbg 0.8032 {\scriptsize $\pm 0.0073$} & 0.9224 {\scriptsize $\pm 0.0084$}\\				
     \cmidrule{2-5}
    \method{ST} & 0.8251 & 0.8601 & \cbg \textbf{0.8284} {\scriptsize $\pm 0.0117$} & 0.9405 {\scriptsize $\pm 0.0085$}\\

    \bottomrule
\end{tabular}
\end{center}
\end{table}

\subsection{Does it matter how the global information is obtained?}
\label{subsec:globalhow}
The most straightforward way to inject global features into the network is by single-cell feature vector aggregation and combining the obtained global vector with each single-cell feature vector for classification. We evaluate max and mean aggregation (\method{MLP-max}, \method{MLP-mean}) as well as using a learned query vector (\method{MLP-pma}). Further, we evaluate \method{PointNet}, a pioneering architecture for 3D point cloud classification and segmentation. Table~\ref{tbl:directglobal} shows no real difference in performance using mean or max aggregation and only a minor performance increase when using a learned query vector. \method{PointNet} performs better, yet the hidden dimensions are much higher (128 and 1024 for single-event and global feature vectors, respectively) than $32$ as used throughout our experiments. When increasing the feature dimension for the single-event vectors to 1024 as well in \method{PointNet-adapted} to remove the increased focus on global information from the standard \method{PointNet} architecture, the model slightly outperforms the current \ac{SOTA} \method{ST} relying on self-attention. This is interesting given the simplicity of \method{PointNet}. However, note that the feature dimension in \method{PointNet-adapted} is 32 times higher than in \method{ST}. 

Further, self-attention can be seen as another way of directly obtaining global information since no local spatial structure is imposed. Table~\ref{tbl:directglobal} shows a clear benefit using feature vectors sampled directly from the \ac{FCM} sample (\method{ST-FPS}) rather than learned query vectors as inducing points. One explanation is that relying more on the sample at hand helps to generalize between patient- or sample-specific shifts and variations (Table~\ref{tbl:ours_gen} supports this interpretation). Further, to compare with a full-range self-attention method, we look at the results obtained with the \method{reluFormer}. Although this approach solely relies on the sample at hand and has all sample-specific information, i.e., the entire sequence length, the performance compared to \method{ST} only improves from $F_1=0.8284$ to $F_1=0.8313$. The \method{reluFormer} is missing the non-linearity of the softmax function in the self-attention operation, which could explain this result. To rule out that the performance increase of \method{ST-FPS} solely comes from using more inducing points, $\approx150$ compared to $16$, we train the \method{ST} with $150$ inducing points and denote this experiment as \method{ST-150I}. We can see, however, that this worsens the results, meaning that the performance increase of \method{ST-FPS} does stain from sampling features vectors of the sample as inducing points. An explanation here could be that sampling the inducing points directly from the data reduces the parameters to be learned, which is beneficial in low data regimes, often the case in \ac{MRD} detection of pediatric leukemia.

Finally, we can answer this question with \textit{yes}, it matters how the global information is obtained, where self-attention that relies solely on the sample at hand and uses a non-linearity to calculate the attention matrices performs best.

\begin{table}
\caption{Results for the methods learning global features in a direct manner.}
\label{tbl:directglobal}
\setlength{\tabcolsep}{8pt}
\begin{center}
\begin{tabular}{clcccc}
    \toprule
    &Method &  $p$ & $r$ & avg $F_1$ & med $F_1$ \\
    \midrule
    \multirow{5}{*}{\rotatebox[origin=c]{90}{\makecell{single-cell\\feature vec. agg.}}}
    
    &\method{MLP-max} &0.7855 & 0.8749 & \cbg 0.8015 {\scriptsize $\pm 0.0044$} & 0.9179 {\scriptsize $\pm 0.0098$}\\		
    \cmidrule{3-6}
    &\method{MLP-mean} & 0.7887 & 0.8748 & \cbg 0.8041 {\scriptsize $\pm 0.0071$}& 0.9203 {\scriptsize $\pm 0.0087$}\\

    \cmidrule{3-6}
    &\method{MLP-pma} & 0.7912 & 0.8755 & \cbg 0.8058 {\scriptsize $\pm 0.0026$}& 0.9228 {\scriptsize $\pm 0.0061$}\\ 		

    \cmidrule{3-6}
    &\method{PointNet} & 0.8027 & 0.8686 & \cbg 0.8117 {\scriptsize $\pm 0.0045$}& 0.9155 {\scriptsize $\pm 0.0073$}\\

    \cmidrule{3-6}
    &\method{PointNet-adapted} & 0.8191 & 0.8792 & \cbg \textbf{0.83} {\scriptsize $\pm 0.0037$} & 0.9437 {\scriptsize $\pm 0.006$}\\

    \midrule
    \multirow{4}{*}{\rotatebox[origin=c]{90}{\makecell{self-\\attention}}}
     &\method{ST-150I} & 0.817 & 0.8597 & \cbg 0.8211 {\scriptsize $\pm 0.012$}& 0.9346 {\scriptsize $\pm 0.008$}\\		

    \cmidrule{3-6}
    &\method{ST} & 0.8251 & 0.8601 & \cbg 0.8284 {\scriptsize $\pm 0.0117$} & 0.9405 {\scriptsize $\pm 0.0085$} \\
    \cmidrule{3-6}
    &\method{reluFormer} & 0.8298 & 0.867 & \cbg 0.8313 {\scriptsize $\pm 0.0059$} & 0.9466 {\scriptsize $\pm 0.0029$} \\						

    \cmidrule{3-6}
    &\method{ST-FPS} & 0.8332 & 0.8636 & \cbg \textbf{0.8369}{\scriptsize $\pm 0.0076$} & 0.9454 {\scriptsize $\pm 0.0063$}\\

    \bottomrule
\end{tabular}
\end{center}
\end{table}

\subsection{Is it beneficial to explicitly learn local features?}
\label{subsec:local_benefits}
To answer this question, we look at local neighborhood aggregation methods and find that \ac{GNN}s are a good fit for \ac{FCM} samples. The graph for each sample, e.g., a sample's local neighborhoods, can be constructed using the full sample-specific features (marker panels). Since marker panels can vary from sample to sample, this is an easy way of incorporating sample-specific information through the structure of the spatial relations, which has to be dismissed entirely in the other methods since all models assessed expect the same set of input features for each sample (see Section \ref{subsec:meth_ours} and Table \ref{tbl:gnn_features}).

We look at three main \ac{GNN} types, \method{GCN}, \method{GAT}, and \method{GIN}, where the latter outperforms \method{GCN} and \method{GAT}. The performance of \method{GAT} and \method{GCN} are similar with incrementally better median $F_1$ score of \method{GAT}. Hence, we conduct all following experiments involving \ac{GNN} with those two. All reach similar results to our baseline \method{ST}, with \method{GIN} even slightly outperforming. Although those methods do not directly learn global features, we assume that this is because with $k=10$, we have connections between every cluster present in the \ac{FCM} sample, and using four layers means that the receptive field of each node (cell) is 400. The local neighborhoods are thus highly overlapping, and information can flow globally. Results for \method{GAT-ASAP}, \method{GIN-ASAP} and only using $k=3$ in the $k$-NN graph construction (\method{GAT-3}, \method{GIN-3}) support this assumption: adding explicit aggregation with \ac{ASAP} has some benefit, yet not significantly, but capping global information flow by using a $k$-NN graph with $k=3$, which results in disconnected clusters within the sample, impairs the results. Although the model cannot learn full-range dependencies, the results are competitive or outperform our baseline \method{ST} indicating that local feature learning is beneficial and that \ac{GNN}s are suitable methods for \ac{FCM} data. 

Based on those results, an initial answer to the question stated is that at least compared to the performance of the current \ac{SOTA}, explicitly modeling local spatial relationships leads to competitive or slightly better results in terms of $F_1$ score and hence modeling long-range dependencies as with self-attention is not strictly necessary to reach those results.

\begin{table}
\caption{Results for methods explicitly learning local features based on \ac{GNN}s.}
\label{tbl:scannet}
\setlength{\tabcolsep}{8pt}
\begin{center}
\begin{tabular}{lcccc}
    \toprule
    Method &  $p$ & $r$ & avg $F_1$ & med $F_1$ \\
    \midrule
     
    \method{GCN} & 0.7827 & 0.8598 & \cbg 0.8288 {\scriptsize $\pm 0.007$} & 0.9405{\scriptsize $\pm 0.0015$}\\

    \cmidrule(l{3pt}r{3pt}){2-5}

     \method{GAT} & 0.7941 & 0.8339 & \cbg 0.8255 {\scriptsize $\pm 0.0063$}& 0.9458 {\scriptsize $\pm 0.0036$}\\	
    \cmidrule(l{3pt}r{3pt}){2-5}

    \method{GIN} & 0.7902 & 0.8486 & \cbg \textbf{0.8317} {\scriptsize $\pm 0.0082$}& 0.9415 {\scriptsize $\pm 0.0045$}\\

     \cmidrule(l{3pt}r{3pt}){1-5}
     \method{GIN-3} & 0.8018 & 0.843 & \cbg 0.813 {\scriptsize $\pm 0.0149$}& 0.9268 {\scriptsize $\pm 0.0138$}\\	
     \cmidrule(l{3pt}r{3pt}){2-5}
     \method{GAT-3} & 0.8104 & 0.8504 & \cbg 0.8147 {\scriptsize $\pm 0.0068$}& 0.9383 {\scriptsize $\pm 0.005$}\\	     
    \cmidrule(l{3pt}r{3pt}){1-5}	

    \method{GAT-ASAP} & 0.7877 & 0.8481 & \cbg 0.8274 {\scriptsize $\pm 0.0086$}& 0.9443 {\scriptsize $\pm 0.0017$}\\	
    \cmidrule(l{3pt}r{3pt}){2-5}
    \method{GIN-ASAP} & 0.7994 & 0.8458 & \cbg \textbf{0.8378} {\scriptsize $\pm 0.0217$}& 0.9457 {\scriptsize $\pm 0.0050$}\\					
    \bottomrule
\end{tabular}
\end{center}
\end{table}

\subsection{\ac{GNN} infused \ac{ST} with \ac{FPS} self-attention}
\label{subsec:meth_ours}
\begin{figure}[t]
  \centering
   \includegraphics[width=1.0\linewidth]{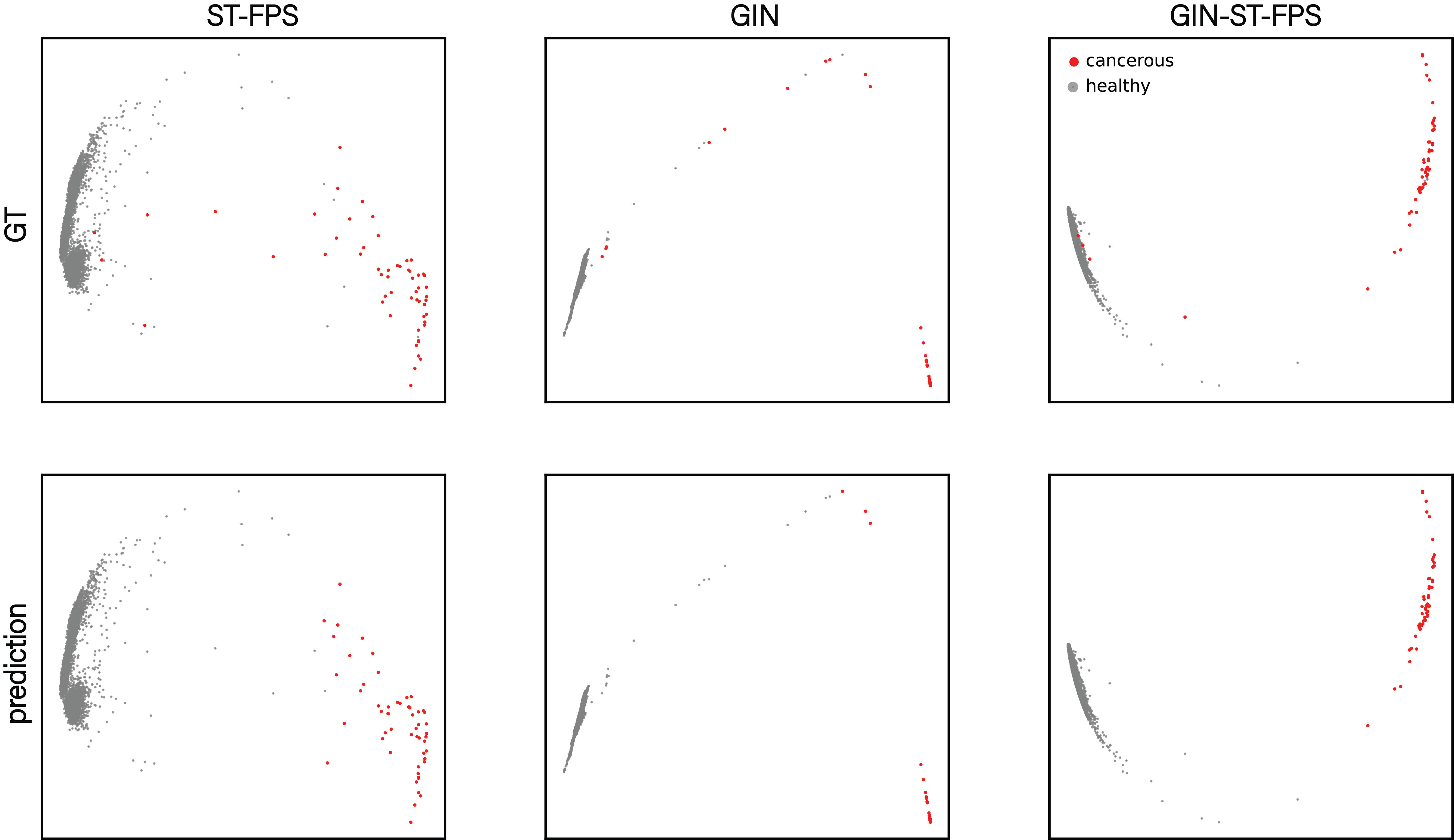}
   \caption{Network features of the last layer of each model before the prediction head (a linear layer) are plotted using PCA with two components. Each model had the same \ac{FCM} sample as input. Healthy cells are denoted in grey, and cancerous cells in red.}
   \label{fig:featureplot}
\end{figure}

Based on the findings above using methods based on either purely learning global or local features, we ask if we can benefit from combining those two strains, similarly to work in \cite{wu2021representing}. 
Table~\ref{tbl:ours} shows that by using \ac{FPS} based \ac{ISAB}s and explicitly learning local features by replacing one layer with a \method{GNN} layer, \method{GAT} or \method{GIN}, we reach new \ac{SOTA} performance on the \textit{Vie} dataset. Further, Table~\ref{tbl:ours_gen} shows that by introducing the adaptations mentioned above, the inter-laboratory generalization ability improves significantly due to relying more on the sample at hand by using \ac{FPS} and the sample's specific spatial local structures. In Figure~\ref{fig:featureplot}, the network features of the last layer before the prediction head, a linear layer, are plotted using PCA with two components of each feature space. The features of \method{ST-FPS}, \method{GIN}, and the combined \method{GIN-ST-FPS} are compared to get insights into how the models might complement each other and create the performance increase. We can see that \method{GIN} compresses the cell types to tight clusters yet struggles with cancer cells that closely adhere to healthy ones. We postulate that this can be traced back to the k-NN graph, where healthy cells might also be strongly connected to cancerous cells on the edge of the different cell population clusters. The \method{ST-FPS} model seems to have a smoother but blurry transition. \method{GIN-ST-FPS}, can use the spatial locality stored in the graph structure while balancing it out by incorporating the context of other cell populations regardless of spacial proximity.

As stated in Section~\ref{subsec:local_benefits}, in the case of varying marker panels between samples, modeling a \ac{FCM} sample as a graph can be beneficial since the information of those sample-specific features can be indirectly incorporated via the graph structure. To analyze if the network can use this information, we remove 3 of the most important features for \ac{b-ALL} detection, CD10, CD19, and CD45, from the input features but keep them for graph construction. Table~\ref{tbl:ours_gen} shows that while the performance drops for all methods, the \method{GAT-ST-FPS} and \method{GIN-ST-FPS} give the best results, confirming our assumption.

Revisiting the question addressed in Section~\ref{subsec:local_benefits}, we can now give a clear \textit{yes} as an answer; explicitly learning local features with \method{GNN} layers, is beneficial, especially when combined with the long-range dependency modeling capabilities of self-attention.

\begin{table}
\caption{Results for our proposed architectural changes.}
\label{tbl:ours}
\setlength{\tabcolsep}{8pt}
\begin{center}
\begin{tabular}{lcccc}
    \toprule
    Method &  $p$ & $r$ & avg $F_1$ & med $F_1$ \\
    \midrule
    \method{ST} & 0.8251 & 0.8601 & \cbg 0.8284 {\scriptsize $\pm 0.0117$} & 0.9405 {\scriptsize $\pm 0.0085$} \\
    \cmidrule(l{3pt}r{3pt}){2-5}
    \method{ST-FPS}& 0.8332 & 0.8636 & \cbg 0.8369 {\scriptsize $\pm 0.0076$} & 0.9454 {\scriptsize $\pm 0.0063$}\\
    \cmidrule(l{3pt}r{3pt}){2-5}
    \method{GAT-ST-FPS} & 0.8242 & 0.8829 & \cbg 0.8465 {\scriptsize $\pm 0.0094$}& 0.9529 {\scriptsize $\pm 0.0043$}\\		
		
    \cmidrule(l{3pt}r{3pt}){2-5}
    \method{GIN-ST-FPS} & 0.8335 & 0.8775 & \cbg \textbf{0.8665} {\scriptsize $\pm 0.0083$}& 0.9561 {\scriptsize $\pm 0.0044$}\\

    \bottomrule
\end{tabular}
\end{center}
\end{table}

\begin{table}
\caption{Results testing generalization ability to datasets of other laboratories.}
\label{tbl:ours_gen}
\setlength{\tabcolsep}{6pt}
\begin{center}
\begin{tabular}{lcccc}
    \toprule
    & \multicolumn{2}{c}{\textit{Bln}} & \multicolumn{2}{c}{\textit{Bue}} \\ 
    Method &  avg $F_1$ & med $F_1$ & avg $F_1$ & med $F_1$  \\
    \midrule
    \method{ST} &  \cbg 0.6089 {\scriptsize $\pm 0.0577	$}& 0.7225 {\scriptsize $\pm 0.0972$}& \cbg 0.7274 {\scriptsize $\pm 0.0214$} & 0.9246 {\scriptsize $\pm 0.0122$} \\				
	
    \cmidrule(l{3pt}r{3pt}){2-5}

    \method{ST-FPS}  & \cbg \textbf{0.7265} {\scriptsize $\pm 0.0275$}& 0.9052 {\scriptsize $\pm 0.0060$}& \cbg 0.7939 {\scriptsize $\pm 0.0164$}& 0.9445{\scriptsize $\pm 0.0095$} \\

    \cmidrule(l{3pt}r{3pt}){2-5}
    \method{GAT-ST-FPS} &  \cbg 0.7136 {\scriptsize $\pm 0.0587$}& 0.8952 {\scriptsize $\pm 0.0488$}& \cbg 0.7802 {\scriptsize $\pm 0.0375$} & 0.9519 {\scriptsize $\pm 0.0156$}\\ 			

    \cmidrule(l{3pt}r{3pt}){2-5}
    \method{GIN-ST-FPS} &  \cbg 0.7046 {\scriptsize $\pm 0.0501$}& 0.9064 {\scriptsize $\pm 0.0406$} & \cbg \textbf{0.8356} {\scriptsize $\pm 0.0111$}& 0.9628 {\scriptsize $\pm 0.0025$}\\

    \bottomrule

\end{tabular}
\end{center}
\end{table}

\begin{table}
\caption{Results showing the model's capability of using implicit information stored in the graph structure. The three most important features for \ac{b-ALL} detection, CD10, CD19, and CD45, were removed from the input features (node features) but kept for graph construction.}
\label{tbl:gnn_features}
\setlength{\tabcolsep}{8pt}
\begin{center}
\begin{tabular}{lcc}
    \toprule
    Method  & avg $F_1$ & med $F_1$ \\
    \midrule
    \method{ST} &  \cbg 0.3462 {\scriptsize $\pm 0.0170$}& 0.1452 {\scriptsize $\pm 0.0565$}\\

    \cmidrule(l{3pt}r{3pt}){2-3}
    \method{ST-FPS}  &  \cbg 0.3668 {\scriptsize $\pm 0.0142$}& 0.2023 {\scriptsize $\pm 0.0554$}\\

    \cmidrule(l{3pt}r{3pt}){2-3}
    \method{GAT-ST-FPS}  & \cbg 0.4343 {\scriptsize $\pm 0.0086$}& 0.3943 {\scriptsize $\pm 0.0285$}\\ 	

    \cmidrule(l{3pt}r{3pt}){2-3}
    \method{GIN-ST-FPS}  & \cbg \textbf{0.4545	{\scriptsize $\pm 0.0113$}} & 0.4647 {\scriptsize $\pm 0.0303$}\\
    \bottomrule
\end{tabular}
\end{center}
\end{table}

\section{Conclusion}
This paper presents an evaluation of different \ac{DL} methods for \ac{FCM} data processing in the problem setting of automated \ac{MRD} detection. Several methods divided into global and local feature learning methods are evaluated, and based on the findings, two adaptations to the current \ac{SOTA} model are proposed. The evaluation shows that modeling long-range dependencies is essential for automated \ac{MRD} detection, where self-attention based on \ac{FPS} performs best. Further, methods based solely on local feature learning can perform similarly and, in some cases, even outperform self-attention-based methods given overlapping local receptive fields. Using feature vectors sampled from the sample at hand by \ac{FPS} instead of learned query vectors combined with introducing a local feature learning layer complement each other and result in a new \ac{SOTA} performance and better inter-laboratory generalization abilities tested on publicly available datasets. 
While the \ac{DL} methods evaluated for automated \ac{FCM} processing in this paper cover a wide range of model types, it is by far not an exhaustive evaluation of possible architectures; extending this work by, e.g., drawing inspiration from 3D point cloud processing for semantic segmentation is thus an interesting topic for future work.


%
%
%
 \bibliographystyle{splncs04}
 \bibliography{ref}

\begin{thebibliography}{10}
\providecommand{\url}[1]{\texttt{#1}}
\providecommand{\urlprefix}{URL }
\providecommand{\doi}[1]{https://doi.org/#1}

\bibitem{abdelaal2019LDA}
Abdelaal, T., van Unen, V., Höllt, T., Koning, F., Reinders, M.J., Mahfouz, A.: Predicting cell populations in single cell mass cytometry data. Cytometry Part A  \textbf{95}(7),  769--781 (2019)

\bibitem{Arvaniti2017cellcnn}
Arvaniti, E., Claassen, M.: Sensitive detection of rare disease-associated cell subsets via representation learning. Nature Communications  \textbf{8}(14825),  2041--1723 (2017)

\bibitem{becht2019dimensionality}
Becht, E., McInnes, L., Healy, J., Dutertre, C.A., Kwok, I.W., Ng, L.G., Ginhoux, F., Newell, E.W.: Dimensionality reduction for visualizing single-cell data using umap. Nature biotechnology  \textbf{37}(1),  38--44 (2019)

\bibitem{bruggner2014citrus}
Bruggner, R.V., Bodenmiller, B., Dill, D.L., Tibshirani, R.J., Nolan, G.P.: Automated identification of stratifying signatures in cellular subpopulations. Proceedings of the National Academy of Sciences  \textbf{111}(26),  E2770--E2777 (2014)

\bibitem{campana2010minimal}
Campana, D.: Minimal residual disease in acute lymphoblastic leukemia. Hematology 2010, the American Society of Hematology Education Program Book  \textbf{2010}(1),  7--12 (2010)

\bibitem{cheung2021currenttrends}
Cheung, M., Campbell, J.J., Whitby, L., Thomas, R.J., Braybrook, J., Petzing, J.: Current trends in flow cytometry automated data analysis software. Cytometry Part A pp. 1--15 (2021)

\bibitem{dash2020deep}
Dash, S., Acharya, B.R., Mittal, M., Abraham, A., Kelemen, A.: Deep learning techniques for biomedical and health informatics. Springer (2020)

\bibitem{dworzak2008standardization}
Dworzak, M.N., Gaipa, G., Ratei, R., Veltroni, M., Schumich, A., Maglia, O., Karawajew, L., Benetello, A., P{\"o}tschger, U., Husak, Z., et~al.: Standardization of flow cytometric minimal residual disease evaluation in acute lymphoblastic leukemia: Multicentric assessment is feasible. Cytometry Part B: Clinical Cytometry: The Journal of the International Society for Analytical Cytology  \textbf{74}(6),  331--340 (2008)

\bibitem{guo2020deep}
Guo, Y., Wang, H., Hu, Q., Liu, H., Liu, L., Bennamoun, M.: Deep learning for 3d point clouds: A survey. IEEE transactions on pattern analysis and machine intelligence  \textbf{43}(12),  4338--4364 (2020)

\bibitem{hu2022application}
Hu, Z., Bhattacharya, S., Butte, A.J.: Application of machine learning for cytometry data. Frontiers in immunology  \textbf{12},  787574 (2022)

\bibitem{kipf2016semi}
Kipf, T.N., Welling, M.: Semi-supervised classification with graph convolutional networks. arXiv preprint arXiv:1609.02907  (2016)

\bibitem{kowarsch2022xai}
Kowarsch, F., Weijler, L., W\"{o}dlinger, M., Reiter, M., Maurer-Granofszky, M., Schumich, A., Sajaroff, E.O., Groeneveld-Krentz, S., Rossi, J.G., Karawajew, L., Ratei, R., Dworzak, M.N.: Towards self-explainable transformers for cell classification in flow cytometry data. In: Interpretability of Machine Intelligence in Medical Image Computing: 5th International Workshop, IMIMIC 2022, Held in Conjunction with MICCAI 2022, Singapore, Singapore, September 22, 2022, Proceedings. p. 22–32. Springer-Verlag, Berlin, Heidelberg (2022)

\bibitem{Lee2017ACDC}
Lee, H.C., Kosoy, R., Becker, C., Kidd, B.: Automated cell type discovery and classification through knowledge transfer. Bioinformatics (Oxford, England)  \textbf{33} (01 2017)

\bibitem{lee2019set}
Lee, J., Lee, Y., Kim, J., Kosiorek, A., Choi, S., Teh, Y.W.: Set transformer: A framework for attention-based permutation-invariant neural networks. In: International Conference on Machine Learning. pp. 3744--3753. PMLR (2019)

\bibitem{Levine2015phenograph}
Levine, J.H., Simonds, E.F., Bendall, S.C., Davis, K.L., Amir, E.a.D., Tadmor, M.D., Litvin, O., Fienberg, H.G., Jager, A., Zunder, E.R., Finck, R., Gedman, A.L., Radtke, I., Downing, J.R., Pe'er, D., Nolan, G.P.: Data-driven phenotypic dissection of aml reveals progenitor-like cells that correlate with prognosis. Cell  \textbf{162}(1),  184–--197 (2015)

\bibitem{li2017gating}
Li, H., Shaham, U., Stanton, K.P., Yao, Y., Montgomery, R.R., Kluger, Y.: Gating mass cytometry data by deep learning. Bioinformatics  \textbf{33}(21),  3423--3430 (2017)

\bibitem{licandro2018wgan}
Licandro, R., Schlegl, T., Reiter, M., Diem, M., Dworzak, M., Schumich, A., Langs, G., Kampel, M.: Wgan latent space embeddings for blast identification in childhood acute myeloid leukaemia. In: 2018 24th International Conference on Pattern Recognition (ICPR). pp. 3868--3873. IEEE (2018)

\bibitem{mckinnon2018flow}
McKinnon, K.M.: Flow cytometry: an overview. Current protocols in immunology  \textbf{120}(1), ~5--1 (2018)

\bibitem{Ni2016SVM}
Ni, W., Hu, B., Zheng, C., Tong, Y., Wang, L., Li, Q.q., Tong, X., Han, Y.: Automated analysis of acute myeloid leukemia minimal residual disease using a support vector machine. Oncotarget  \textbf{7}(44),  71915--71921 (2016)

\bibitem{qi2017pointnet}
Qi, C.R., Su, H., Mo, K., Guibas, L.J.: Pointnet: Deep learning on point sets for 3d classification and segmentation. In: Proceedings of the IEEE conference on computer vision and pattern recognition. pp. 652--660 (2017)

\bibitem{qin2022cosformer}
Qin, Z., Sun, W., Deng, H., Li, D., Wei, Y., Lv, B., Yan, J., Kong, L., Zhong, Y.: cosformer: Rethinking softmax in attention. arXiv preprint arXiv:2202.08791  (2022)

\bibitem{ranjan2020asap}
Ranjan, E., Sanyal, S., Talukdar, P.: Asap: Adaptive structure aware pooling for learning hierarchical graph representations. In: Proceedings of the AAAI conference on artificial intelligence. vol.~34, pp. 5470--5477 (2020)

\bibitem{reiter2019automated}
Reiter, M., Diem, M., Schumich, A., Maurer-Granofszky, M., Karawajew, L., Rossi, J.G., Ratei, R., Groeneveld-Krentz, S., Sajaroff, E.O., Suhendra, S., et~al.: Automated flow cytometric mrd assessment in childhood acute b-lymphoblastic leukemia using supervised machine learning. Cytometry Part A  \textbf{95}(9),  966--975 (2019)

\bibitem{rota2016role}
Rota, P., Kleber, F., Reiter, M., Groeneveld-Krentz, S., Kampel, M.: The role of machine learning in medical data analysis. a case study: Flow cytometry. In: VISIGRAPP (3: VISAPP). pp. 305--312 (2016)

\bibitem{suffian2022machine}
Suffian, M., Montagna, S., Bogliolo, A., Ortolani, C., Papa, S., D’Atri, M., et~al.: Machine learning for automated gating of flow cytometry data. In: CEUR WORKSHOP PROCEEDINGS. vol.~3307, pp. 47--56. Sun SITE Central Europe, RWTH Aachen University (2022)

\bibitem{testi2019outcome}
Testi, A.M., Attarbaschi, A., Valsecchi, M.G., M{\"o}ricke, A., Cario, G., Niggli, F., Silvestri, D., Bader, P., Kuhlen, M., Parasole, R., et~al.: Outcome of adolescent patients with acute lymphoblastic leukaemia aged 10--14 years as compared with those aged 15--17 years: Long-term results of 1094 patients of the aieop-bfm all 2000 study. European Journal of Cancer  \textbf{122},  61--71 (2019)

\bibitem{vaswani2017attention}
Vaswani, A., Shazeer, N., Parmar, N., Uszkoreit, J., Jones, L., Gomez, A.N., Kaiser, {\L}., Polosukhin, I.: Attention is all you need. In: Advances in neural information processing systems. pp. 5998--6008 (2017)

\bibitem{velickovic2018gat}
Veličković, P., Cucurull, G., Casanova, A., Romero, A., Liò, P., Bengio, Y.: Graph attention networks. In: International Conference on Learning Representations (2018)

\bibitem{Weber2019cdiffcyt}
Weber, L.M., Nowicka, M., Soneson, C., Robinson, M.D.: diffcyt: Differential discovery in high-dimensional cytometry via high-resolution clustering. Communications Biology  \textbf{2}(183),  2399--3642 (2019)

\bibitem{weijler2020umaprf}
Weijler, L., Diem, M., Reiter, M., Maurer-Granofszky, M.: Detecting rare cell populations in flow cytometry data using umap. In: 2020 25th International Conference on Pattern Recognition (ICPR). pp. 4903--4909 (2021)

\bibitem{weijler2024fate}
Weijler, L., Kowarsch, F., Reiter, M., Hermosilla, P., Maurer-Granofszky, M., Dworzak, M.: Fate: Feature-agnostic transformer-based encoder for learning generalized embedding spaces in flow cytometry data. In: Proceedings of the IEEE/CVF Winter Conference on Applications of Computer Vision. pp. 7956--7964 (2024)

\bibitem{weijler2022umaphdbscan}
Weijler, L., Kowarsch, F., Wödlinger, M., Reiter, M., Maurer-Granofszky, M., Schumich, A., Dworzak, M.N.: Umap based anomaly detection for minimal residual disease quantification within acute myeloid leukemia. Cancers  \textbf{14}(4) (2022)

\bibitem{wodlinger2021automated}
Wodlinger, M., Reiter, M., Weijler, L., Maurer-Granofszky, M., Schumich, A., Groeneveld-Krentz, S., Ratei, R., Karawajew, L., Sajaroff, E., Rossi, J., Dworzak, M.N.: Automated identification of cell populations in flow cytometry data with transformers. Computers in Biology and Medicine p. 105314 (2022)

\bibitem{wu2021representing}
Wu, Z., Jain, P., Wright, M., Mirhoseini, A., Gonzalez, J.E., Stoica, I.: Representing long-range context for graph neural networks with global attention. Advances in Neural Information Processing Systems  \textbf{34},  13266--13279 (2021)

\bibitem{xu2018powerful}
Xu, K., Hu, W., Leskovec, J., Jegelka, S.: How powerful are graph neural networks? In: International Conference on Learning Representations (2019)

\end{thebibliography}

\end{document}